# ONBOARD MARINE ANOMALY DETECTION ON ΦSAT-2: FROM SIMULATION-BASED DEVELOPMENT TO IN-ORBIT DEMONSTRATION

**Clotilde Szywala, Thomas Goudemant, Marjorie Bellizzi, Benjamin Francesconi, Adrien Girard**

***Institut de Recherche Technologique Saint Exupéry***
*240 Rue Evariste Galois*
*06410 Biot, France*
*Email: firstname.lastname@irt-saintexupery.com*

## ABSTRACT

Onboard Artificial Intelligence can improve responsiveness and bandwidth efficiency of Earth Observation systems by processing data directly on the satellite. This paper presents the experience gained from the development, onboard integration, and post-launch adaptation of a lightweight marine anomaly detection pipeline deployed on the European Space Agency's Φsat-2 mission. The application combines sea segmentation, self-supervised feature encoding of marine regions, generic anomaly detection based on deviations from a normal sea state, and optional characterization of selected anomaly types. Before launch, the pipeline was trained and validated on simulated Φsat-2 imagery to assess algorithmic performance and compatibility with resource-constrained onboard hardware. After integration and functional validation in the mission environment, early experiments on real Φsat-2 acquisitions revealed a significant mismatch between simulated and in-orbit data. The pipeline was therefore retrained on real Level-1 imagery using an improved annotation strategy to better handle ambiguous marine regions, substantially enhancing performance. Beyond demonstrating the onboard feasibility of the application, the Φsat-2 experience highlights the importance of robust annotation strategies and sensor-aware design, and shows that simulation-based development is valuable for pre-flight risk reduction, while reliable scientific validation requires representative in-orbit data and should be clearly distinguished from functional validation.

## 1. INTRODUCTION

Conventional Earth Observation (EO) processing chains largely rely on the downlink of large volumes of data followed by ground-based analysis, which can introduce latency and consume limited communication bandwidth [1]. Onboard Artificial Intelligence (AI) can improve system responsiveness by processing data directly on the satellite to filter irrelevant acquisitions, prioritize observations, or provide early insights before transmission [2]. However, onboard AI must operate under stringent resource and integration constraints, particularly in terms of computing capability, memory footprint, and compatibility with mission execution environments [3, 4]. Representative training data are also limited: actual sensor imagery is unavailable before launch, while rare events may remain poorly represented and difficult to annotate even after deployment [5]. Models are therefore often developed using mission-representative simulated data and must subsequently be reassessed and, when necessary, adapted using real in-orbit imagery.

Marine environmental monitoring is a representative use case, as harmful algal blooms, oil spills, and sediment plumes can have major ecological and socioeconomic impacts [6, 7, 8]. Existing approaches often target these phenomena separately, using spectral band combinations [9, 10] or supervised models for predefined event types [11]. Such approaches may be sensor- and data-dependent, limiting their generalization to other missions and rare or unseen events. Unsupervised and self-supervised anomaly detection offer a more generic, less label-dependent alternative [12, 13], but few studies address the complete onboard chain, including constrained execution, mission integration, and validation on real in-orbit data [14].

In this context, we developed a lightweight and modular pipeline for onboard marine anomaly detection. Rather than learning each anomaly type explicitly, it models the normal sea state and identifies deviations from this learned normality, while an optional characterization stage provides more specific information on known phenomena. Combining compact self-supervised representations, domain-driven spectral features, and classical machine-learning models, the pipeline balances genericity, data frugality, and compatibility with constrained onboard hardware.

This paper focuses on the deployment of this pipeline on the European Space Agency's (ESA) Φsat-2 mission, a 6-unit (6U) CubeSat launched in 2024 and equipped with an eight-band Visible and Near-Infrared (VNIR) camera with a 4.75 m Ground Sampling Distance (GSD) and an Intel Movidius Myriad 2 accelerator [15, 2]. Although the pipeline was also adapted to other demonstration contexts such as IMAGIN-e and Kanyini [15], it reached its most advanced integration and validation stage on Φsat-2, providing the most comprehensive operational experience. This paper therefore covers the main steps of this process: development and pre-flight validation on simulated Φsat-2 imagery, onboard integration and functional validation, analysis of the differences observed between simulated data and the first in-orbit acquisitions, retraining on real Φsat-2 data, and the resulting lessons learned for future onboard AI missions.

## 2. PIPELINE ARCHITECTURE AND DESIGN CHOICES

Our anomaly detection pipeline builds on the architecture introduced in [4] and subsequently refined in [16]. As shown in Fig. 1, it follows a modular structure composed of five sequential stages:

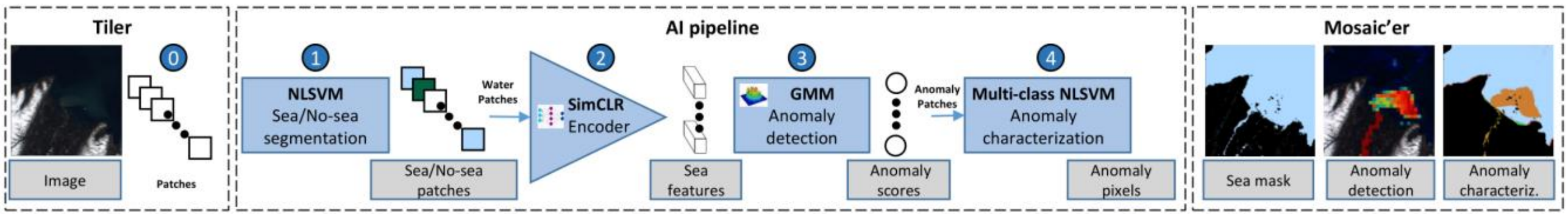

Fig. 1. Architecture of the anomaly detection pipeline: The Tiler (0) splits the input image into patches, which are processed by the AI pipeline. Sea patches are filtered (1), encoded into a latent space (2), and analyzed for anomalies (3). Anomalies are segmented (4) to identify their type. The Mosaic'er then recombines the patches for visualization.

0. *Tiling:* The input image is split into non-overlapping fixed-size patches (e.g., 32 × 32 pixels) for local analysis.
1. *Sea/no-sea segmentation:* A Non-Linear Support Vector Machine (NLSVM) performs per-pixel sea/no-sea segmentation within each patch. Patches are retained for subsequent processing only when the proportion of sea pixels exceeds a predefined threshold, in order to restrict anomaly detection to marine areas. This model relies on selected spectral bands, band ratios, and indices, injecting domain knowledge into the classification process and making it physically interpretable, data-efficient, and computationally lightweight.
2. *Sea patch encoding*: Each sea patch is projected into a compact latent space using a lightweight convolutional encoder trained with the self-supervised SimCLR method [17]. This approach captures relevant spatial and spectral information while reducing data dimensionality and limiting the need for manual annotations.
3. *Anomaly detection:* A Gaussian Mixture Model (GMM) models the distribution of normal sea patches in the encoder latent space using several Gaussian components, and assigns an anomaly score to each patch based on its deviation from this learned normality. This enables generic and data-efficient detection of various marine anomalies without requiring explicit examples of each anomaly type.
4. *Anomaly characterization:* When a patch is classified as anomalous, with an anomaly score exceeding a predefined threshold, a multi-class NLSVM, composed of three parallel NLSVMs—sea vs. oil, sea vs. algae, and sea vs. sediments—can perform per-pixel segmentation within the patch into four classes: sea, oil, algae, and sediments. Like the sea/no-sea segmentation stage, this model relies on selected spectral features.

This architecture was designed to balance genericity, algorithmic performance, data frugality, and onboard feasibility. By combining lightweight, independently trained components rather than relying on a single end-to-end model, the pipeline remains modular, interpretable, and adaptable to mission-specific constraints. Individual stages can therefore be fine-tuned, retrained or reconfigured, and the spectral configuration adjusted, without redesigning the entire system.
Operationally, this pipeline is intended to support two key applications: prioritization of images with high anomaly scores for downlink or ground analysis, and alerting when significant marine events, such as oil spills, are detected.

## 3. DEVELOPMENT AND VALIDATION ON SIMULATED DATA

At the time the Φsat-2 pipeline was designed and trained, the satellite was not yet in orbit and no image acquired by its multispectral camera was available. The first development phase was therefore conducted on simulated datasets representative of the Φsat-2 sensing conditions, with the objective of reducing technical and integration risks before onboard deployment. This phase enabled the initial training and validation of the models, the assessment of hardware feasibility, and the definition of a first performance baseline to be reassessed later using real in-orbit data (Section 5).

### 3.1. Training on Simulated Data

#### *3.1.1. Simulated Dataset Construction and Overview*

Synthetic images representative of Φsat-2 sensing conditions were generated as Level-1C (L1C) products, corresponding to Top-Of-Atmosphere (TOA) reflectance values, from Sentinel-2 data using the ESA simulator [5]. Two datasets were constructed: a marine normality dataset, manually annotated with sea/no-sea masks, and an anomaly dataset built from scenes corresponding to documented oil spills, algal blooms, and sediment plumes, manually annotated with land, sea, cloud, and anomaly masks. Oil spills were retained as an exploratory target despite expected spectral limitations (Section 3.2). The dataset construction and simulation processes have been described in previous publications [4, 16, 15].
The annotation of the anomaly dataset nevertheless raised several practical difficulties. In particular, defining clear boundaries between normal sea and large, diffuse anomalies, such as algal blooms or sediment plumes, proved challenging. Some anomalies were also difficult to identify reliably from the imagery alone, despite the use of scientific literature to select acquisition dates and locations.

#### *3.1.2. Training Configuration*

The complete pipeline was implemented in Python. The sea/no-sea and multi-class NLSVMs were trained in a supervised manner, with features (bands, indices, and ratios) selected to maximize class discrimination. The encoder was trained in a self-supervised manner on 32 × 32-pixel patches extracted from the marine normality dataset using SimCLR [17], projecting each patch into a 16-dimensional latent space. Finally, the GMM was trained in a weakly supervised manner on normal sea patches to model the distribution of normal marine conditions in this latent space.
Several training configurations and alternative anomaly detection methods were evaluated during development. The detailed training procedures and comparative evaluations have been reported in previous publications [4, 16, 15].

### 3.2. Algorithmic Performance

Each pipeline stage was evaluated independently on the ground using scenes from the marine normality and anomaly datasets, assuming perfect operation of the preceding stage: anomaly detection was evaluated on ground-truth sea patches, and anomaly characterization on ground-truth anomalous patches. Fig. 2 shows an example of sediment detection.
The performance results (Table 1) indicate a good overall capability of the pipeline. Sea/no-sea segmentation achieves an F1-score of almost 94%. Algae and sediment anomalies are also detected and characterized satisfactorily, with detection F1-scores above 60% and characterization recalls above 87%. In contrast, oil spill detection and characterization are more challenging, with both precision and recall significantly lower for detection, and recall below 70% for characterization.
This lower performance was not unexpected, as oil spills had already been identified as a challenging target for Φsat-2: their discrimination can benefit from Short-Wave Infrared (SWIR) information, used by several Sentinel-2-based approaches [18] but unavailable on Φsat-2's eight-band VNIR sensor. Oil/sea discrimination therefore relies on limited spectral contrast and is more sensitive to factors such as sunglint, sea state, and oil thickness, which likely contributed to the observed performance. In addition, oil spills appeared difficult to identify reliably in the simulated imagery (Section 3.1.1), likely introducing uncertainty in the ground-truth masks. These interpretations are supported by the weak apparent oil/sea separability observed in the data, which may reflect both sensor limitations and annotation uncertainty.
More generally, the measured performance should be interpreted in light of uncertainties in the manual annotations. In particular, errors may arise from overly strict ground-truth masks in areas where the boundary between anomalies and sea is ambiguous (Section 3.1.1). This limitation is especially relevant for anomaly characterization, where apparent false positives may correspond to weakly affected or transitional areas that were not included in the anomaly masks.
Overall, these results provided a useful pre-flight baseline for Φsat-2. They validated the relevance of the modular architecture and confirmed that algae and sediment anomalies could be handled satisfactorily on simulated data. At the same time, they highlighted two points to be addressed during the in-orbit phase: the need for more reliable annotations in ambiguous transition areas, and the limited robustness of oil spill detection with the available spectral configuration.

### 3.3. Hardware Performance

Prior to onboard deployment, the pipeline was assessed against the constraints of low-power onboard processing in order to verify its hardware efficiency. The models were executed on platforms representative of the Φsat-2 onboard system: the encoder was evaluated on a Myriad 2 accelerator, and the NLSVMs and GMM on an ARM Cortex-A53 processor.
The results (Table 2) confirm that the NLSVMs and GMM have limited impact on processing time, while the encoder, despite its lightweight architecture of approximately one million parameters, is the main computational bottleneck. Based on these latencies, the processing time for a typical 4,096 × 4,096-pixel acquisition is estimated at about five minutes, which would allow several acquisitions to be processed per day, depending on the available onboard processing windows.
Overall, this pre-flight assessment validated the feasibility of the modular and frugal pipeline for Φsat-2 deployment, while highlighting the importance of model sizing for constrained onboard execution.

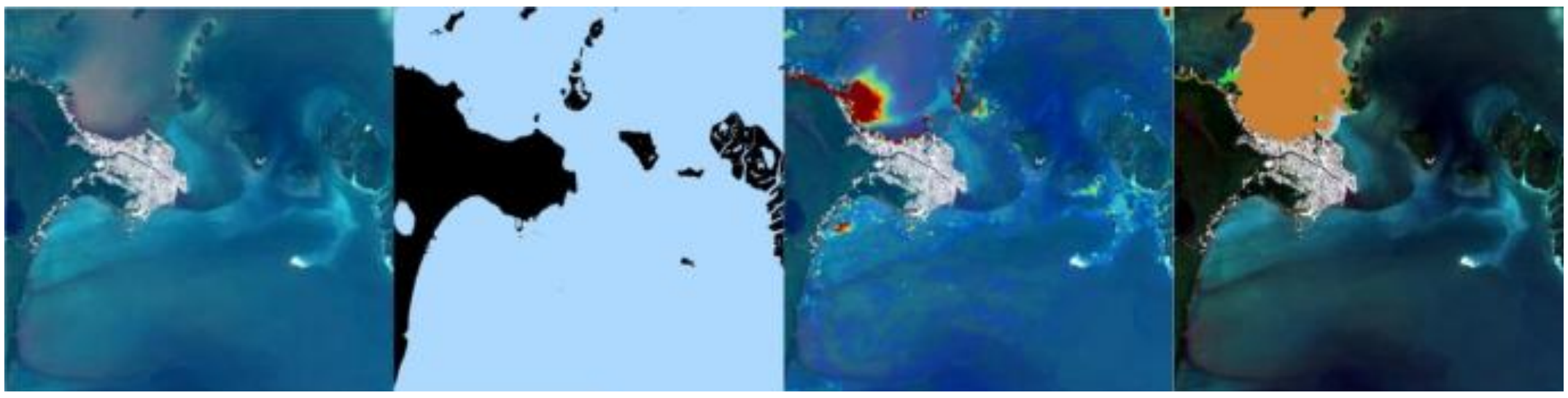

Fig. 2. Outputs of the pipeline for a simulated image: RGB visualization, sea/no-sea segmentation map (blue: sea, black: no-sea), anomaly detection heatmap on ground-truth sea patches (red: scores close to 1, blue: scores close to 0), anomaly characterization map on ground-truth anomalous patches (blue: sea, green: algae, brown: sediments).

Table 1. Algorithmic performance of each pipeline stage, evaluated independently on simulated data.

*(a) Sea/no-sea segmentation (sea/no-sea NLSVM)*

| Number of pixels | Precision | Recall | F1-score |
|---|---|---|---|
| $1.0103 \times 10^9$ (66 images) | 95.7% | 88.8% | 93.9% |

*(b) Anomaly detection (encoder + GMM)*

| Anomaly type | Number of patches | Precision | Recall | F1-score |
|---|---|---|---|---|
| Algae | 5,167 (2 images) | 82.6% | 63.9% | 72.1% |
| Oil | 4,063 (10 images) | 4.0% | 2.8% | 3.3% |
| Sediments | 7,206 (5 images) | 66.5% | 55.1% | 60.3% |

*(c) Anomaly characterization (multi-class NLSVM)*

| Anomaly type | Number of pixels | Recall |
|---|---|---|
| Algae | $4.9 \times 10^6$ (2 images) | 99.4% |
| Oil | $2.2 \times 10^6$ (7 images) | 68.2% |
| Sediments | $10.2 \times 10^6$ (6 images) | 87.4% |

Table 2. Latency and throughput of the pipeline models on hardware representative of the Φsat-2 onboard processing system. Encoder results use a batch size of 128 patches, identified as optimal for hardware-accelerated inference.

| AI model | Target | Latency ($\mu$s/patch) | Throughput (patches/s) |
|---|---|---|---|
| Sea/no-sea NLSVM | ARM Cortex-A53 | $2.0 \times 10^2$ | $5.0 \times 10^3$ |
| Encoder | Myriad 2 | $1.7 \times 10^4$ | $6.0 \times 10^1$ |
| GMM | ARM Cortex-A53 | $2.9 \times 10^1$ | $3.4 \times 10^4$ |
| Multi-class NLSVM | ARM Cortex-A53 | $1.1 \times 10^3$ | $8.9 \times 10^2$ |

## 4. ONBOARD INTEGRATION AND EARLY IN-ORBIT FEEDBACK

The development and validation activities presented in Section 3 reduced the main algorithmic and hardware risks of deploying the pipeline, which was then delivered to the Φsat-2 mission team for onboard integration. This phase aimed to verify execution of the complete processing chain in a mission-representative environment before confirming onboard execution. This functional validation process consisted of comparing outputs generated in the mission environment with reference outputs generated on ground servers. Scientific relevance was not assessed at this stage, as the first Φsat-2 acquisitions revealed a significant mismatch between the simulated data and the available in-orbit products.

### 4.1. Onboard Deployment and Functional Validation

The pipeline was ported from Python to Java and first integrated and validated in the Φsat-2 FlatSat, a ground-based system-level bench representative of the satellite interfaces and flight software stack. The objective was to verify that it could be launched and run to completion in a mission-representative environment before onboard deployment.
Functional validation consisted of running the pipeline on a Φsat-2 acquisition in the mission environment and comparing its anomaly detection and characterization outputs with ground-reference outputs generated from the same input data.
For anomaly detection, patch-level anomaly scores, ranging from 0 to 1, were compared between the two executions. The outputs were considered consistent, with a mean absolute error of $6 \times 10^{-4}$. This is illustrated in Fig. 3, which shows visually consistent outputs and an absolute error map indicating that the remaining differences are sparse and localized.
For anomaly characterization, outputs were compared at pixel level after setting patches with anomaly scores below 0.3 to zero in both outputs. Over 16,777,216 pixels, 6,560 disagreements were observed, corresponding to a disagreement rate of 0.04%. These discrepancies were likely caused by rounding effects around the threshold value and were sufficiently limited to consider the outputs consistent.
Overall, this validation step confirmed correct integration and execution of the pipeline in the mission environment, with residual differences attributable to numerical implementation effects. The pipeline was then deployed and successfully executed onboard Φsat-2 through the CogniSAT execution environment, where it processed several in-orbit acquisitions.

### 4.2. Early Insights from First Φsat-2 Data

In addition to the functional validation, early Φsat-2 acquisitions were used to qualitatively assess how the pipeline trained on simulated data (Section 3.1) behaved on real mission data. The pipeline was applied to two representative images: one Level-0 (L0) and one Level-1 (L1) product.
The real Φsat-2 L1 products available for these experiments were stored as 12-bit images with values in the [0, 4,095] range and had undergone absolute radiometric correction. However, radiance-to-TOA-reflectance conversion had not been applied, in contrast to the simulated training images, which were generated as L1C products [5], corresponding to TOA reflectance values typically in the [0, 1] range. Real-data inference was therefore performed after dividing input values by 4,095 to approximate the numerical range used during training.
Results on the L1 image are shown in Fig. 4. Sea/no-sea segmentation remains qualitatively satisfactory, while anomaly detection and characterization produce many false positives, with large areas of normal sea being interpreted as anomalous and mostly characterized as sediments. Similar behavior was observed on the L0 image. These results reveal a marked mismatch between the simulated training data and the real Φsat-2 products available for these experiments.
Several factors may contribute to the observed differences. First, the simple normalization applied to the real L1 products was insufficient to make them radiometrically consistent with the simulated L1C reflectance data used for training. A more appropriate preprocessing would likely require converting the real L1 products to TOA reflectance before inference. Second, the simulated training data were generated from Sentinel-2 acquisitions (Section 3.1.1), which cannot fully reproduce the spatial and spectral characteristics of the Φsat-2 sensor, notably its 4.75 m GSD compared with 10–60 m for Sentinel-2. Such sensor-related differences may contribute to a domain gap between simulated and real in-orbit data. Overall, these first experiments highlight that, although the pipeline executed correctly onboard, its detection behavior is limited by the mismatch between the simulated training domain and the actual Φsat-2 sensor and processing chain. Fine-tuning or retraining on in-orbit data was therefore required before assessing scientific performance.

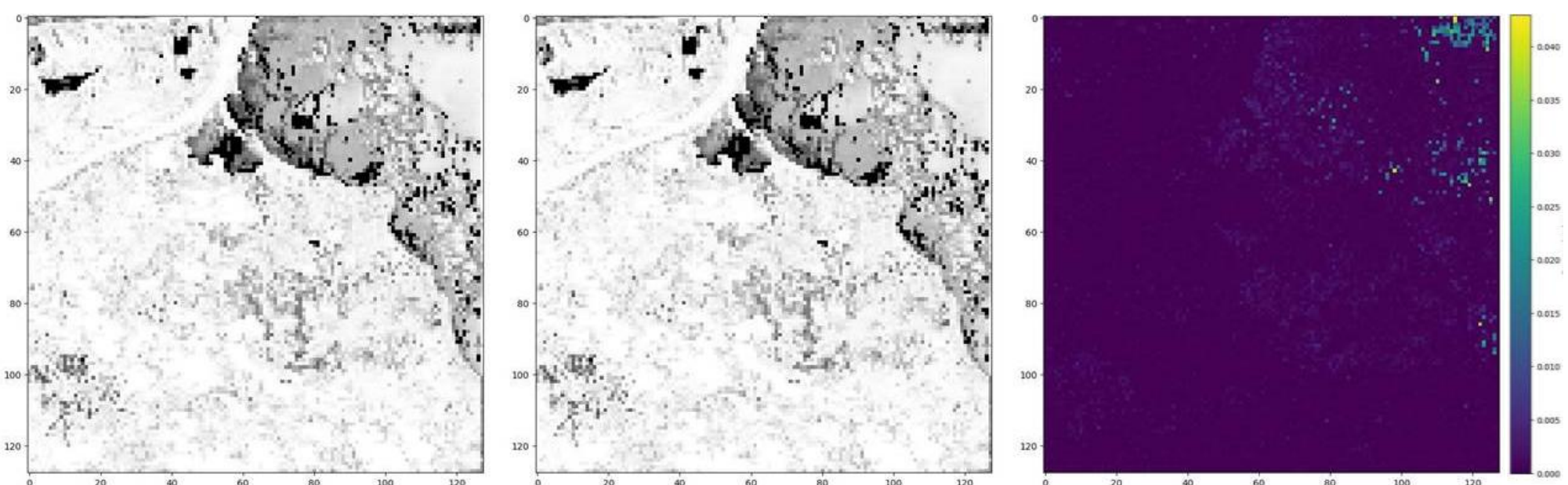

Fig. 3. Comparison of anomaly detection outputs on a Φsat-2 L0 image: ground-reference output (left), mission-environment output (middle), and absolute error map (right).

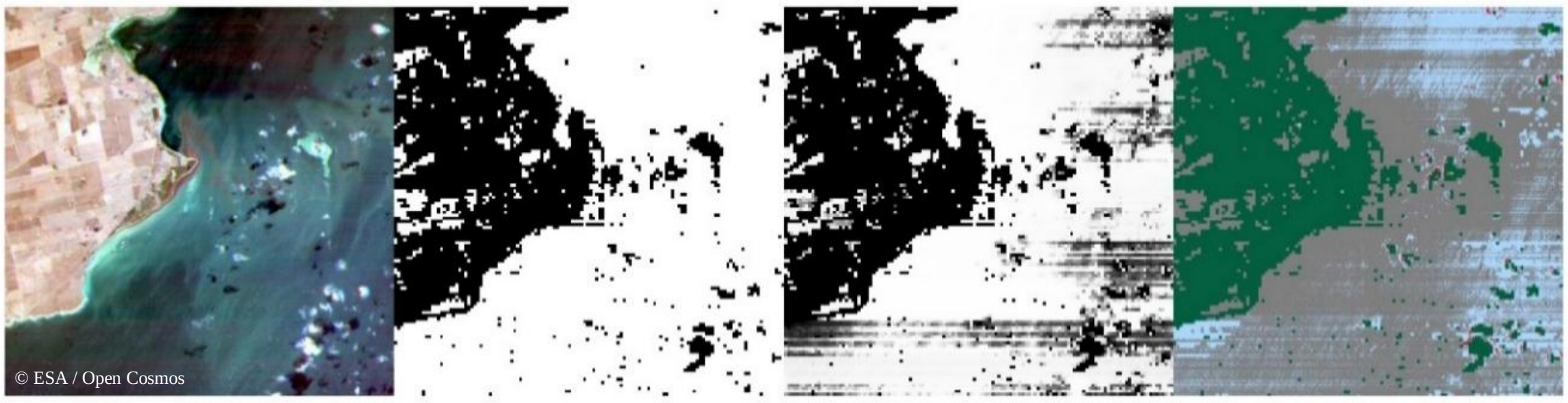

Fig. 4. Outputs of the pipeline for a Φsat-2 L1 image: RGB visualization, sea/no-sea segmentation map (white: sea, black: no-sea), anomaly detection heatmap on segmented sea patches (white: scores close to 1, black: scores close to 0), anomaly characterization map on detected anomalous patches (blue: sea, red: oil, brown: sediments, dark green: unprocessed).

## 5. ADAPTATION TO REAL MISSION DATA

As anticipated during development and confirmed by the first Φsat-2 acquisitions presented in Section 4.2, a significant mismatch was observed between simulated and in-orbit imagery. Although fine-tuning was initially considered, with real Φsat-2 L1 data becoming available, retraining was chosen as a direct and pragmatic approach to adapt the pipeline to the actual sensor and processing chain. The lessons learned from the simulated-data validation phase, in particular regarding annotation uncertainty and class ambiguities, were used to improve both the construction of the real dataset and the associated training strategy, with the objective of reducing evaluation bias. This new phase aimed to improve and robustify the algorithmic performance of the pipeline on real Φsat-2 data, before validating its scientific relevance in orbit.

### 5.1. Retraining on Real Data

#### *5.1.1. Real Dataset Construction and Annotation*

To support the construction of a real Φsat-2 marine dataset, we provided the mission teams responsible for acquisition planning and tasking with a list of geographical points of interest targeting scenes useful for pipeline retraining and evaluation. These points were selected from scientific literature and press reviews, and mainly covered coastal areas where relevant marine phenomena, such as algal blooms and sediment plumes, were likely to occur. Priority was given to estuarine regions, where suspended sediments are transported and often trapped [19]. Such scenes are particularly valuable because a single acquisition may contain several classes required by the pipeline: land areas, sometimes affected by clouds, clear water areas representative of marine normality, and potentially anomalous water areas.

An initial real Φsat-2 dataset was then built from 4,096 × 4,096-pixel L1 images. It included 17 images defining normal marine conditions (11 training, 6 testing) and 7 images containing sediment or algae anomalies (5 training, 2 testing).

Oil spill samples were excluded because of the limited performance they showed on simulated data (Section 3.2), likely due to annotation uncertainty and weak oil/sea separability with the available Φsat-2 bands. Representative in-orbit oil samples were also more difficult to acquire than recurrent coastal phenomena such as sediment plumes or algal blooms.

The annotation strategy was revised based on the lessons learned during the simulated-data phase. In addition to the land, cloud, sea, sediments, and algae classes, three ambiguous classes were introduced: “sea or algae”, “sea or sediments”, and “algae or sediments”. The objective was to better handle uncertain class boundaries, especially in transition areas between clear water and diffuse anomalies, or in regions where sediments and algae may coexist or be difficult to distinguish, in order to reduce annotation bias. Their use during training and evaluation is described in Section 5.1.2. To further improve annotation reliability, preliminary masks were generated automatically using spectral indices to support the identification of anomaly and transition areas. These masks were then manually completed and refined by image-processing experts. Examples of final ground-truth masks, including ambiguous classes, are shown in Fig. 5.

#### *5.1.2. Pipeline Retraining*

The pipeline models were retrained on the real Φsat-2 dataset using a training configuration largely consistent with the one applied during the simulated-data phase (Section 3.1.2). However, two adaptations were introduced to account for the lessons learned from the simulated-data validation phase: the oil spill class was excluded from the retraining, and the ambiguous classes (Section 5.1.1) were handled in a specific manner during training and evaluation.

These ambiguous classes were not treated as additional detection targets, but were introduced to avoid forcing uncertain pixels into a single reference class during training and evaluation, thereby reducing bias at uncertain class boundaries. For anomaly detection, ambiguous pixels were excluded from training and metric computation, ensuring that the GMM was trained only on unambiguous normal sea patches and that uncertain transition areas did not affect the detection scores.

For anomaly characterization, each ambiguous label was converted into a multi-hot target representing the compatible classes; for example, a pixel annotated as “sea or algae” was considered compatible with both sea and algae. The multi-class NLSVM was therefore trained with a sigmoid output layer and weighted binary cross-entropy loss on these multi-hot targets, so that the model could learn from uncertain pixels without forcing a single arbitrary class assignment. During evaluation, a prediction on an ambiguous pixel was considered correct if it matched any compatible class.
Finally, since the real Φsat-2 dataset contained no oil spill samples, this class was not considered during retraining of the multi-class NLSVM. Its architecture was kept unchanged for compatibility with the existing pipeline, but the oil spill output was disabled and not used for prediction.

### 5.2. Updated Algorithmic Performance

Algorithmic performance was evaluated on the ground on both the training and test Φsat-2 image sets. As with simulated data, each stage was assessed independently, assuming perfect operation of the preceding stage: anomaly detection was assessed on ground-truth sea patches, and anomaly characterization on ground-truth anomalous patches.
The results (Table 3) show very strong performance, exceeding that obtained during the simulated-data validation phase (Section 3.2), with F1-scores above 94% for all three processing steps on both datasets. This suggests that the retraining strategy adopted after the first in-orbit feedback (Section 4.2) was effective. In particular, the introduction of ambiguous classes in the annotation masks likely helped reduce annotation bias by better accounting for uncertain transition areas.
However, these results should be interpreted with caution, as the dataset remains limited in size, with only 16 and 8 scenes used for training and testing, respectively. In addition, the introduction of ambiguous classes reduces the penalty associated with physically plausible predictions, but also makes the evaluation protocol more tolerant than a strict single-label assessment, which may contribute to higher measured performance. Further evaluation on a larger and more diverse dataset would therefore be required to more thoroughly assess the generalization capability of the retrained models. Additional fine-tuning iterations may also be needed to consolidate the pipeline’s performance.
A full end-to-end execution of the retrained pipeline was also performed on both the training and test datasets. Intermediate outputs obtained for two Φsat-2 images containing sediment plumes are shown in Fig. 5. In both cases, the final output indicates that sea and sediments are consistently identified, including in transition areas annotated as “sea or sediments”, where the model assigns pixels to either sea or sediments while preserving a coherent transition between both classes. This illustrates the role of ambiguous labels in handling uncertain class boundaries. On the test image, some areas are also characterized as algae, which may correspond either to false positives or to real algal patterns that were not identified during ground-truth creation. This reflects a broader difficulty of the use case: diffuse or mixed marine phenomena can be difficult to identify, so missing annotations do not always guarantee the absence of the phenomenon.
Overall, lessons from the simulated-data phase effectively guided the real-dataset construction and the retraining strategy, leading to a more robust and less biased evaluation. This first iteration significantly improved performance on real data, but should be seen as an initial operational validation that will need to be consolidated using a more diverse dataset.

Table 3. Algorithmic performance of each pipeline stage, retrained and evaluated independently on real Φsat-2 L1 data.

| Model | Training dataset (16 images) | | | Test dataset (8 images) | | |
|---|---|---|---|---|---|---|
| | Precision | Recall | F1-score | Precision | Recall | F1-score |
| Sea/no-sea NLSVM | 99.09% | 98.03% | 98.55% | 99.53% | 89.60% | 94.31% |
| Encoder + GMM | 99.98% | 99.59% | 99.79% | 100.0% | 89.17% | 94.28% |
| Multi-class NLSVM | 99.95% | 99.93% | 99.94% | 100.0% | 94.45% | 97.14% |

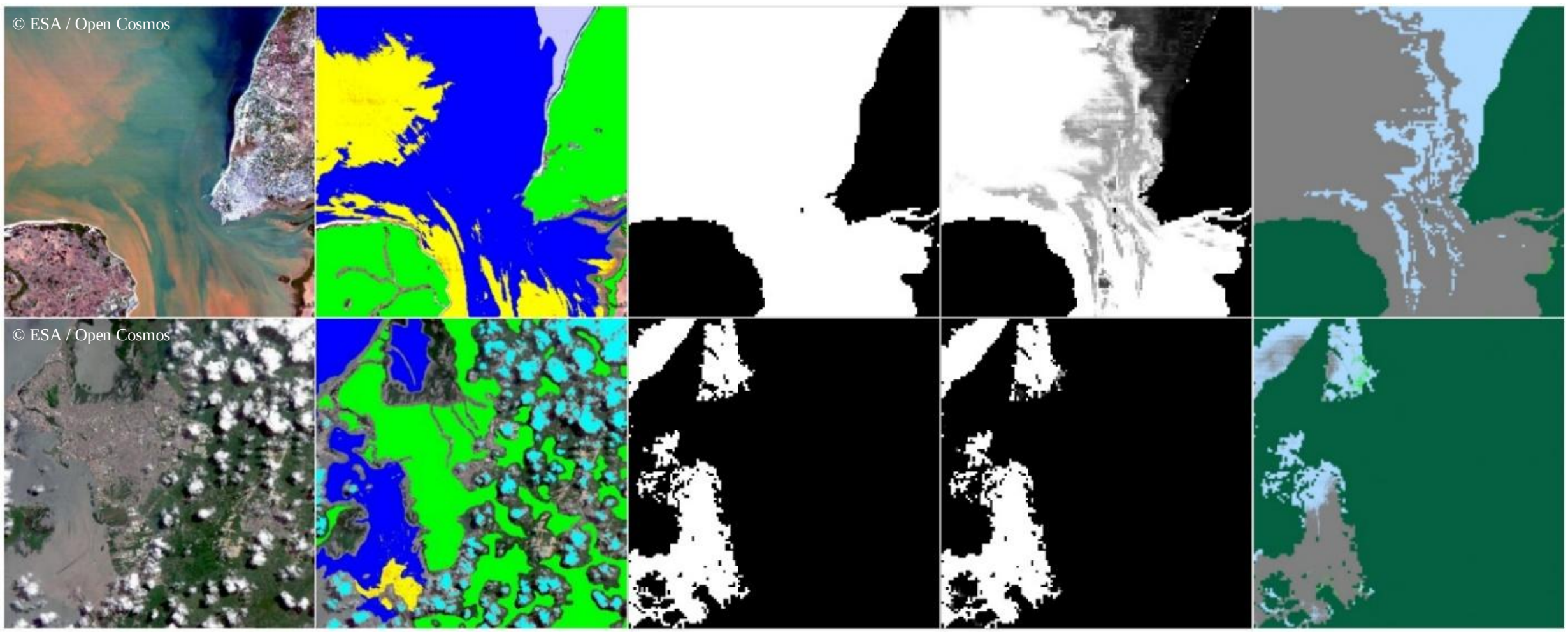

Fig. 5. Outputs of the pipeline for two Φsat-2 L1 images. Image 1: training example; Image 2: test example. Columns show RGB visualizations, ground-truth masks (light blue: sea, blue: sea or sediments, yellow: sediments, orange: algae or sediments, green: land, cyan: clouds), sea/no-sea segmentation maps (white: sea, black: no-sea), anomaly detection heatmaps on segmented sea patches (white: scores close to 1, black: scores close to 0), and anomaly characterization maps on detected anomalous patches (blue: sea, green: algae, brown: sediments, dark green: unprocessed).

## 6. LESSONS LEARNED AND PERSPECTIVES

The Φsat-2 deployment provided a practical end-to-end experience of bringing a lightweight AI pipeline from simulation-based development to onboard validation. Beyond demonstrating onboard feasibility, this experience offered valuable feedback on the methodological, operational, and scientific challenges associated with deploying marine anomaly detection in orbit. These lessons provide a basis for consolidating the in-orbit validation process and for identifying future extensions toward more operational use.

### 6.1. Lessons Learned from the Φsat-2 Deployment

In the absence of real Φsat-2 imagery before launch, simulated data were essential to design, train, optimize, and validate the pipeline before deployment, providing a first performance baseline. However, the first in-orbit acquisitions showed that this pre-flight validation could not guarantee scientific performance on real mission data, because the simulated training data did not fully reproduce the actual Φsat-2 sensor and processing chain (Section 4.2). This highlights the need to treat simulated-data validation as a pre-flight risk-reduction step, rather than as a final scientific validation. It also underlines the importance of separating two validation levels: functional validation, which verifies that the application launches, runs to completion, complies with mission interfaces, and produces outputs consistent with ground references, and scientific validation, which assesses the physical relevance of the detections, their robustness to real scenes, and their ability to generalize.

The experiments also highlighted that diffuse marine phenomena, such as sediment plumes or algal blooms, cannot always be represented by strict single-label masks: in transition areas between clear water and anomalies, or when several phenomena coexist, the ground truth may be intrinsically uncertain. Introducing ambiguous classes better reflected this physical uncertainty and reduced annotation bias by avoiding the arbitrary assignment of uncertain pixels to a single class (Section 5.1). More generally, this experience shows that annotation strategy should be addressed early in the development process, as it affects ground-truth quality, annotation effort, training and evaluation robustness, and measured performance. Future remote-sensing applications involving diffuse or mixed phenomena should therefore consider flexible annotation protocols, such as ambiguous or multi-label classes, soft labels, confidence masks, or exclusion zones, that can be complemented, when possible, by expert validation, in situ measurements, and dedicated remote-sensing tools.

Beyond domain adaptation and annotation uncertainty, scientific performance remains bounded by the spectral information available onboard. The oil spill case is particularly illustrative: with an eight-band VNIR sensor and no SWIR bands, oil/sea separability is limited and highly dependent on acquisition conditions [18] (Section 3.2). This experience therefore provided a useful assessment of the performance limits imposed by the sensor's spectral configuration.

Finally, for this application, the modular architecture proved useful after launch. It made it possible to retrain all models on real mission data, using training subsets and objectives adapted to each stage, while narrowing the first real-data validation to sediment and algae anomalies and incorporating ambiguous labels into the anomaly-characterization training and evaluation strategy, without redesigning the full processing chain (Section 5.1.2). However, training the blocks independently also limits global optimization: errors can propagate between stages, so end-to-end performance may be lower than that measured for each block. A possible alternative for future developments would be to use a more unified model or a multi-block pipeline trained jointly, while preserving sufficient modularity for diagnosis and targeted updates.

### 6.2. Future Work and Perspectives

The first priority is to upload the retrained pipeline to Φsat-2 and to perform an end-to-end scientific in-orbit validation on new acquisitions. This validation should assess the physical relevance of the detections on real scenes including representative areas, such as clear waters, coastal areas, clouds, sediment plumes, and algal blooms, as well as the in-orbit hardware performance of the pipeline, in order to confirm its robustness under operational conditions.

To go further, one perspective is to progressively extend the real Φsat-2 dataset to cover a wider range of acquisition conditions, geographical contexts, and anomaly types, through better-targeted acquisitions supported, when possible, by expert interpretation, in situ information, or external tools. A larger, more diverse, and more representative dataset could support an iterative update strategy: acquisition and annotation of new images, ground-based fine-tuning and validation, upload to the satellite, and renewed in-orbit evaluation. If suitable data become available, oil spill detection could also be reassessed, although this remains secondary compared with the validation of the current algae and sediment use cases.

The modular structure of the pipeline also enables its extension with complementary expert models. As part of the Φsat-2 mission extension, the current pipeline will be retained as the main generic anomaly-detection branch, while an expert model developed and provided by the University of Sannio (Università degli Studi del Sannio) will be integrated in parallel with the anomaly-characterization stage to target specific water-quality parameters.

Beyond Φsat-2, future work could build on the adaptations to IMAGIN-e and Kanyini mentioned in Section 1 by exploring coordinated acquisitions with complementary missions around common marine monitoring scenarios, when orbital conditions allow comparable observations. This would help compare onboard detections across sensors and assess the portability of the solution across platforms. In the longer term, such coordinated acquisitions could also support the investigation of multi-sensor fusion strategies.

## ACKNOWLEDGMENTS

The work presented in this article was carried out within the IRMA project ("Image processing for a Responsive Mission with AI") at IRT Saint Exupéry. The authors thank the industrial and academic partners of the project: Thales Alenia Space, Geo4i, JoliBrain, and Université Côte d'Azur. They also gratefully acknowledge ESA Φ-lab and the Φsat-2 mission team, particularly Open Cosmos and CGI, for enabling the integration and in-orbit demonstration of the marine anomaly detection application. They further thank SmartSat CRC and the Kanyini mission team, as well as Thales Alenia Space in Spain and Microsoft for the IMAGIN-e demonstration, for their contributions to the multi-mission framework in which this work was developed. This work was partly funded by the French France 2030 program.